\documentclass{article}
\usepackage{spconf,amsmath,graphicx,subcaption}
\usepackage{cite}

\usepackage{tikz}
\usepackage{lipsum}

\makeatletter
\newcommand{\printfnsymbol}[1]{%
  \textsuperscript{\@fnsymbol{#1}}%
}
\makeatother

\newcommand\copyrighttext{%
  \footnotesize © 2020 IEEE.  Personal use of this material is permitted.  Permission from IEEE must be obtained for all other uses, in any current or future media, including reprinting/republishing this material for advertising or promotional purposes, creating new collective works, for resale or redistribution to servers or lists, or reuse of any copyrighted component of this work in other works.}
\newcommand\mycopyrightnotice{%
\begin{tikzpicture}[remember picture,overlay]
\node[anchor=north,yshift=-30pt,xshift=0.5pt] at (current page.north) {\fbox{\parbox{\dimexpr\textwidth-\fboxsep-\fboxrule\relax}{\copyrighttext}}};
\end{tikzpicture}%
}

\title{Interpreting medical image classifiers by optimization based counterfactual impact analysis}
\name{David Major\thanks{\printfnsymbol{1}Equal contribution}\printfnsymbol{1}, Dimitrios Lenis\printfnsymbol{1}, Maria Wimmer, Gert Sluiter, Astrid Berg, and Katja B\"uhler \thanks{VRVis is funded by BMVIT, BMDW, Styria, SFG and Vienna Business Agency in the scope of COMET - Competence Centers for Excellent Technologies (854174) which is managed by FFG. Thanks go to our project partner AGFA HealthCare for providing valuable input.}}
\address{VRVis Zentrum f\"ur Virtual Reality und Visualisierung Forschungs-GmbH, Vienna, Austria}
\begin{document}

%
\maketitle

\begin{abstract}
Clinical applicability of automated decision support systems depends on a robust, well-understood classification interpretation. Artificial neural networks while achieving class-leading scores fall short in this regard. 
Therefore, numerous approaches have been proposed that map a salient region of an image to a diagnostic classification. Utilizing heuristic methodology, like blurring and noise, they tend to produce diffuse, sometimes misleading results, hindering their general adoption.

In this work we overcome these issues by presenting a model agnostic saliency mapping framework tailored to medical imaging. We replace heuristic techniques with a strong neighborhood conditioned inpainting approach, which avoids anatomically implausible artefacts. We formulate saliency attribution as a map-quality optimization task, enforcing constrained and focused attributions. Experiments on public mammography data show quantitatively and qualitatively more precise localization and clearer conveying results than existing state-of-the-art methods.

\end{abstract}

\setlength{\fboxrule}{0pt}
\mycopyrightnotice

\section{Introduction}
\label{sec:intro}

Consulting radiologists will routinely back their findings by pinpointing and describing a specific region on a radiograph. Contrary, acting as a highly efficient black box, artificial neural networks (ANN) \cite{litjens2017} fall short of this form of explanation for their predictions. ANNs' high dimensional, nonlinear nature, does not induce a canonical map between derived prediction and input image. Understandably, a plethora of approaches have been presented that try to derive a so called saliency map, that is, a robust mapping between pixel space and prediction class \cite{zintgraf2017,fong2017,dabkowski2017,simonyan2013,zhou2016,shrikumar17,chang2018,petsiuk2018,uzunova2019}. 

\begin{figure}[ht!]
	\begin{center}
		\includegraphics[width=.48\textwidth]{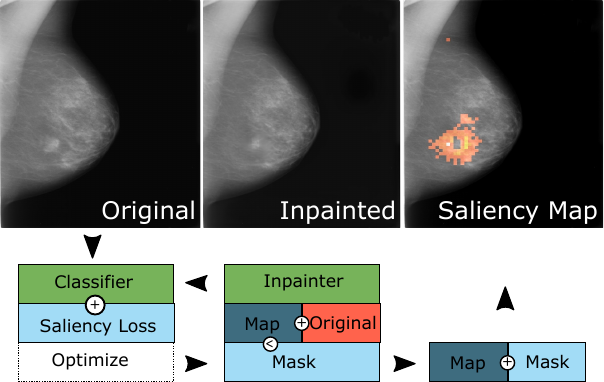}
	\end{center}
	\vspace{-0.5cm}
	\caption{Our saliency mapping framework: (i) classification score of an image is obtained, (ii) a hole mask is generated, inpainted and classified, (iii) saliency loss is computed based on the score difference of original/inpainted images and map quality, (iv) optimization is continued for a fixed number of steps and a result map/mask is derived.}
	\label{fig:overview}
\end{figure}
Most frequently this form of reasoning is based on \emph{local explanations} (LE), i.e. on concrete maps for image-prediction pairs \cite{fong2017,lipton2016}. 
A clinically applicable LE needs to be \emph{informative} for the radiologists, that is, focusing on regions coinciding with medical knowledge \cite{lombrozo2006}.  Moreover, a methodologically sound LE is \emph{faithful} to the classifier, i.e. dependent on architecture, parametrization, and its preconditions like training-set distribution \cite{adebayo2018}.

\emph{Direct approaches} efficiently utilize the assumed analytic nature or the layered architecture of an ANN classifier to derive the desired saliency map for a LE \cite{simonyan2013,zhou2016}. While frequently applied, the obtained results of this class are possibly incomplete, diffuse, hard to interpret, and as recent work shows misleading \cite{zintgraf2017,fong2017,dabkowski2017,shrikumar17,adebayo2018}. Thereby they violate both criteria, informativeness and faithfulness, hindering their general application in medical imaging.

Contrary, \emph{reference based} LE approaches \cite{fong2017} try to mitigate these issues by studying how the given classifier reacts to perturbations of the input image. Using the original input as a reference and marginalizing a dedicated image region's contribution, they estimate this region's effect on the classification score. Solutions mainly vary in the ways this marginalization is achieved. They range from heuristic approaches, e.g. blurring, noise, or graying out \cite{dabkowski2017,fong2017}, over local neighbourhood conditioning \cite{zintgraf2017}, to utilizing strong conditional generative models \cite{chang2018,uzunova2019}. 
These methods address \emph{informativeness}, however, applied to medical images, they introduce noise, possibly pathological indications, anatomical implausible tissue or other adversarial artefacts. By this, they amplify the out-of-distribution problem, similar to an adversarial attack: they expect a meaningful classification result for an image, that is not within the training-set distribution. Hence, they fall short of \emph{faithfulness} for clinical applications.

Marginalization for medical imaging, i.e. the replacement of pathological regions with counterfactual healthy tissue, is being actively explored and addressed by generative adversarial network setups (GAN). Besides promising results, authors report resolution limitations, and the same underlying out-of-distribution issue\cite{bermudez2018,baumgartner2017,becker2019,andermatt2019}.
  
\textbf{Contribution:} We address the open challenge of \emph{faithful and informative} medical black-box classifier interpretation by expanding natural image classifier visualization approaches~\cite{zintgraf2017,dabkowski2017,fong2017}.
We propose a reference based optimization framework tailored to medical images, focusing on the interactions between original and marginalized image classification-scores, and map quality.
To tackle anatomical correctness of marginalization in medical images, partial convolution inpainting \cite{liu2018} is adapted. Hence, instead of a globally acting GAN, we utilize local per-pixel reconstruction without sacrificing global image composition. 
We validate our approach on publicly available mammography data, and show quantitatively and qualitatively more precise localization, and clearer conveying results than existing state-of-the-art methods.

\section{Methods}
\label{sec:methods}

Our goal is to estimate a \emph{faithful} and \emph{informative} saliency map between a medical image and its classification score: given an image, we search for and visually attribute the \emph{specific} pixel-set that contributes towards a confident classification for a fixed class (see Fig.~\ref{fig:overview}). Following \cite{dabkowski2017,zintgraf2017} we formulate the general problem as finding the \emph{smallest deletion region} (SDR) of a class $c$, i.e. the pixel-set whose marginalization w.r.t. the classifier lowers the classification score for $c$. 

\textbf{Image-wise Saliency Mapping}: Informally, we search for the smallest smooth map, that indicates the regions we need to change (inpaint) such that we get a \emph{sufficiently healthy} image able to \emph{fool the classifier}. We formalize the problem as follows: 
Let $I$ denote an image of a domain $\mathcal{I}$ with pixels on a discrete grid $m_1 \times m_2$, $c$ a fixed class, and $f$ a classifier capable of estimating $p(c|I)$, the probability of $c$ for $I$. Also let $M$ denote the saliency mask for image $I$ and class $c$, hence $M \in M^{m_1 \times m_2}(\{0,1\})$. We use \emph{total variation} $tv(M)$ \cite{dabkowski2017}, and \emph{size} $ar(M)$, to measure the mask's shape. Note that \emph{size} here is ambiguous. Experimentally we found dice overlap with regions-of-interest like organ masks to be favourable over the map's average pixel value\cite{dabkowski2017}. With $\odot$ denoting elementwise multiplication, and $ \pi(M)$ the inpainting result of a hole image $I \odot M$, we can define  $ \phi(M) := -1 \cdot \log ( p(c| \pi(M)) ) $ and $\psi(M) := \log (\text{odds}(I))  - \log (\text{odds}(\pi(M))) $, where  $\text{odds}(I) =  \frac{ p(c|I) }{ 1 -  p(c|I)}$. Both, $\phi$ and $\psi$, weigh the new probability of the inpainted image. If we assume class $c$ to denote \emph{pathological}, then healthy images, and large score differences will be favoured. 
With this preparation we define our desired optimization function as
\begin{equation*}
\mathcal{L}(M) := \lambda_1 \cdot (\phi(M) + \psi(M)) + \lambda_2 \cdot tv(M) + \lambda_3 \cdot ar(M)  
\end{equation*}
where $\lambda_i \in {\rm I\!R}$ are regularization parameters, and search for $\arg\min_{M} \; \mathcal{L}(M)$. There are two collaborating parts in $\mathcal{L}$. The first term enforces the class probability to drop, the latter two emphasize an informative mask. Focusing on medical images, $\mathcal{L}$ directly solves the SDR task, thereby minimizing medically implausible and adversarial artefacts caused by inpainting of large classifier-neutral image regions, as observable in \cite{liu2018,dabkowski2017,fong2017}. 
The optimization problem is solved by local search through stochastic gradient descent, starting from a regular grid initialization. By design, no restrictions are applied on the classifier $f$. For optimization we relax the mask's domain to $M^{m_1 \times m_2}([0,1])$, and threshold at $\theta \in (0,1)$. 

\textbf{Image Inpainting with Partial Convolutions}: For marginalization, we want to emphasize local context, while still considering global joint region interaction, and thereby favor a globally sound anatomy. Therefore, we adapt the U-Net like architecture of \cite{liu2018}, capable of handling masks with irregular shapes, fitting our optimization requirements for pathological regions of different sizes and shapes. The chosen architecture consists of eight partial convolution layers on both encoding and decoding parts. It takes an image with holes $I \odot M$ and the hole mask $M$ as an input, and outputs the inpainted image $\pi(M)$.
The partial convolution layers insert only the convolution result of the current sliding convolution-window when image information is present. The convolution filter $W$ is applied on the features $X$
using the binary mask $M$ and yields new features $x'$ the following way:

\begin{figure}[ht!]
\centering
\vspace{-0.725cm}
\begin{minipage}{0.5\linewidth}
\raggedright
\includegraphics[width=6.7cm,height=6.7cm]{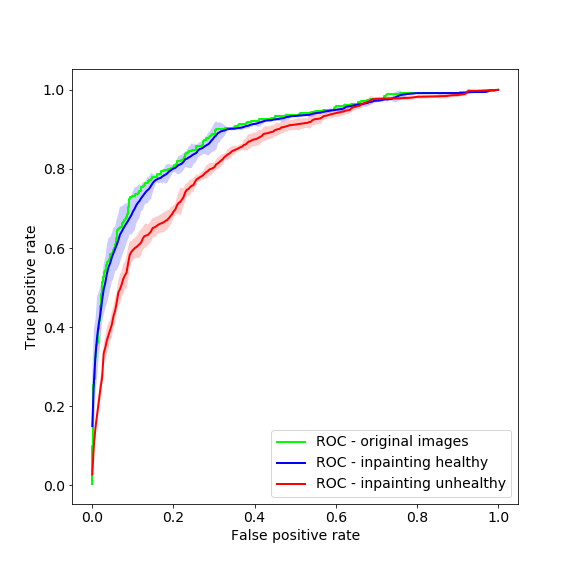}
\end{minipage}%
\begin{minipage}{0.5\linewidth}
\raggedleft
\includegraphics[width=2.5cm,height=2.5cm]{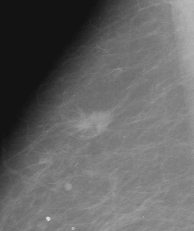}
\par \medskip \vfill
\includegraphics[width=2.5cm,height=2.5cm]{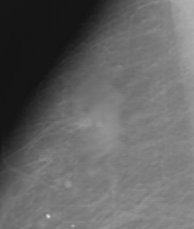}
\end{minipage}
\vspace{-0.5cm}
\caption{Left: Comparison of classifier performance without inpainting (green), with inpainting in healthy tissue (blue) and in mass tissue (red) over 10 random runs (shadowed). Right: Original image with mass (top), inpainted with replaced healthy texture (bottom).}
\label{fig:inpainter_performance}
\end{figure}

\vspace{-0.35cm}
\begin{equation*}
x' = \begin{cases}
W^{T}(X\odot M)\frac{1}{sum(M)}+b, & \text{if} \;sum(M)>0 \\
0, & \text{otherwise}
\end{cases}
\end{equation*}
where $b$ is the bias term. The convolution operation is scaled by $\frac{1}{sum(M)}$ according to the amount of information available in the current sliding window.
Moreover a new mask is passed to the next layer which is updated by setting its values to $1$ in the sliding window if $sum(M)>0$.
The authors of \cite{liu2018} propose to train the network with a loss function concentrating on both per-pixel reconstruction performance of the hole/non-hole regions
and on overall appearance of the image. To improve the overall appearance a perceptual loss and a style loss are applied which match images in a mapped feature space.
Total variation is used as a last loss component to ensure a smooth transition between hole regions and present image regions.

\begin{table}
\begin{center}
\begin{tabular}{|c|c|c|c|}
\hline
P & \textit{$D_{ours}$} & \textit{$D_{cam}$} & \textit{$D_{sal}$} \\
\hline
50 & \textbf{137.1$\pm$69.5} & 200.1$\pm$65.3 & 164.2$\pm$36.0   \\
\hline
75 & \textbf{137.1$\pm$69.5} & 182.1$\pm$78.9 & 162.9$\pm$36.9 \\
\hline
90 & \textbf{137.1$\pm$69.5} & 144.1$\pm$89.3 & 166.3$\pm$44.0  \\
\hline
Ablation & 163.3$\pm$35.9 & - & - \\
\hline
\hline
P & \textit{$H_{ours}$} & \textit{$H_{cam}$} & \textit{$H_{sal}$}\\
\hline
50 &  \textbf{227.1$\pm$82.1} & 359.3$\pm$83.9 & 371.3$\pm$61.5  \\
\hline
75 & \textbf{227.1$\pm$82.1} & 308.5$\pm$103.4 & 350.3$\pm$63.1\\
\hline
90 & \textbf{227.1$\pm$82.1} & 232.2$\pm$121 & 328.9$\pm$64.5  \\
\hline
Ablation & 328.1$\pm$69.5 & - & -  \\
\hline
\end{tabular}
\end{center}
\vspace{-0.5cm}
\caption{Results of the saliency mapping on test data with masses for different percentiles $P$:
$D...$ average Euclidean, $H...$ average Hausdorff distances,
between result mask connected components and ground-truth annotations (in pixels).}
\label{tab:results_optimization}
\end{table}

\begin{table}
\begin{tabular}{|c|c|c|c|c|c|c|c|c|}
\hline
P & \textit{$A_{ours}$} & \textit{$A_{cam}$} & \textit{$A_{sal}$} & \textit{$O_{cam}$} & \textit{$O_{sal}$} \\
\hline
50  &\textbf{.02$\pm$.02} & .43$\pm$.10& .50$\pm$.0 & .60$\pm$ .33 & .86$\pm$.14  \\
\hline
75 &\textbf{.02$\pm$.02} & .24$\pm$.03& .25$\pm$.0 & .44$\pm$ .32 & .64$\pm$.22  \\
\hline
90 &\textbf{.02$\pm$.02} & .10$\pm$.0& .10$\pm$.0 & .23$\pm$ .25 & .39$\pm$.23  \\
\hline
\end{tabular}
\vspace{-0.2cm}
\caption{For three threshold-levels, chosen by percentiles of the output-distribution of \emph{CAM} resp. \emph{SAL}:
$A$: Average overlaps between attributions and breast area; $O$: Overlap between our attribution and  \emph{CAM} resp. \emph{SAL}.} 
\label{tab:results_overlap}
\end{table}

\begin{figure*}[htb!]
\centering
  \begin{subfigure}[b]{.23\linewidth}
    \centering
    \includegraphics[width=.99\textwidth,trim={0 10 0 21},clip]{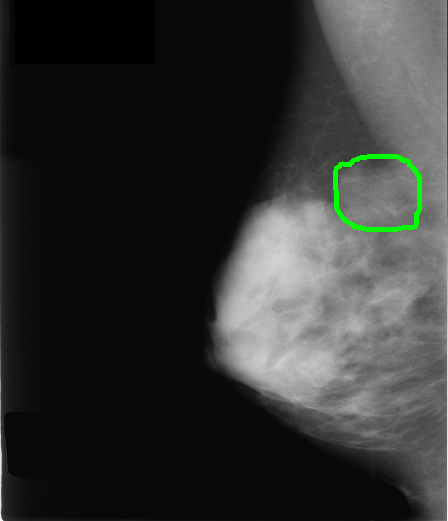}
  \end{subfigure}%
  \begin{subfigure}[b]{.23\linewidth}
    \centering
    \includegraphics[width=.99\textwidth,trim={0 10 0 20},clip]{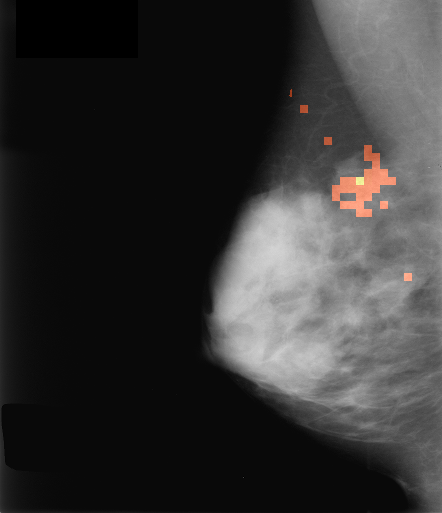}
  \end{subfigure}%
  \begin{subfigure}[b]{.23\linewidth}
    \centering
    \includegraphics[width=.99\textwidth,trim={0 10 0 21},clip]{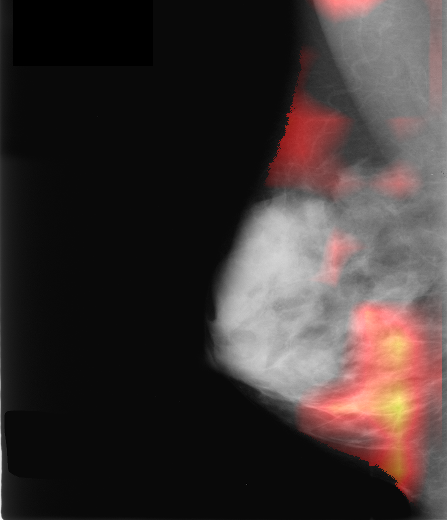}
  \end{subfigure}%
	\begin{subfigure}[b]{.23\linewidth}
    \centering
    \includegraphics[width=.99\textwidth,trim={0 10 0 22},clip]{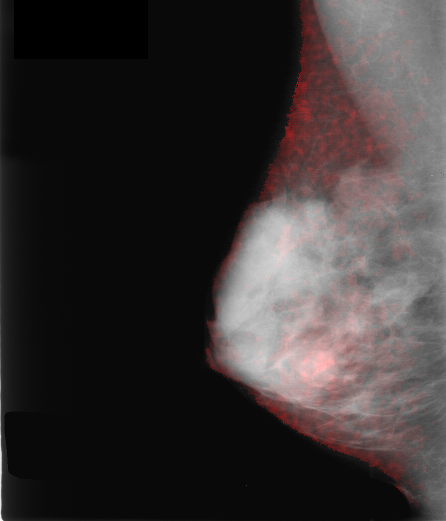}
  \end{subfigure}%
	
	\begin{subfigure}[b]{.23\linewidth}
    \centering
    \includegraphics[width=.99\textwidth,trim={0 20 0 10},clip]{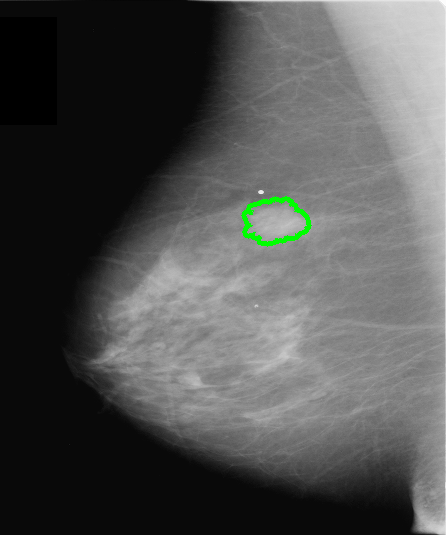}
    \caption{}\label{fig:1e}
  \end{subfigure}%
  \begin{subfigure}[b]{.23\linewidth}
    \centering
    \includegraphics[width=.99\textwidth,trim={0 21 0 10},clip]{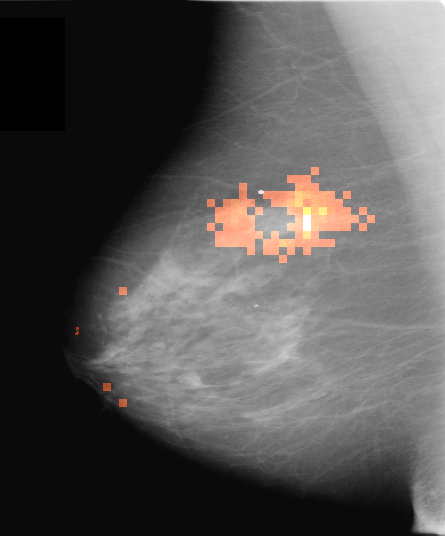}
    \caption{}\label{fig:1f}
  \end{subfigure}%
  \begin{subfigure}[b]{.23\linewidth}
    \centering
    \includegraphics[width=.99\textwidth,trim={0 20 0 10},clip]{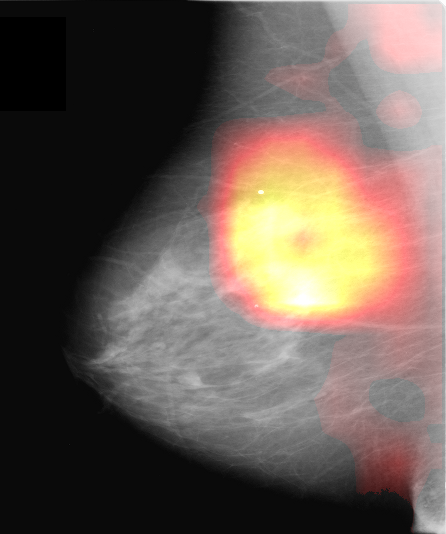}
    \caption{}\label{fig:1g}
  \end{subfigure}%
	\begin{subfigure}[b]{.23\linewidth}
    \centering
    \includegraphics[width=.99\textwidth,trim={0 20 0 10},clip]{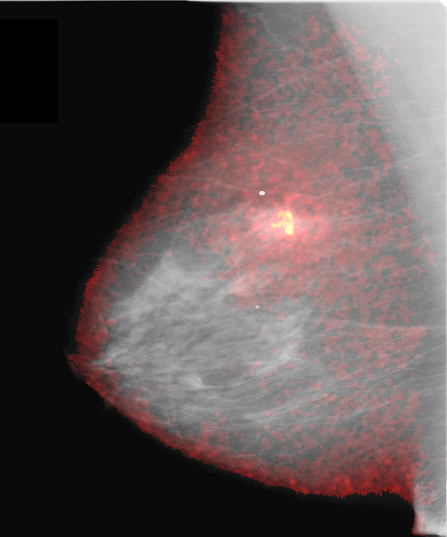}
    \caption{}\label{fig:1h}
  \end{subfigure}%
\vspace{-0.3cm}
\caption{Result saliency heatmaps: (a) depicts the original image overlayed with mass annotation contours, (b) shows results of our saliency mapping framework. (c) is the CAM of the classifier and (d) visualizes SAL, both thresholded by 50th-percentile.} 
\label{fig:results}
\end{figure*}

\section{Experimental Setup}
\label{sec:methods}

\textbf{Dataset:} In this work the Database for Screening Mammography (DDSM)~\cite{heath2000} and the Curated Breast Imaging Subset of DDSM (CBIS-DDSM)~\cite{lee2016} were used,
downsampled to a resolution of 576x448 pixels. 
Data was split into 1231 scans containing masses and 2000 healthy samples for training, and into 334 mass and 778 healthy scans for testing.
Scans with masses contain pixel-wise ground-truth annotation (GT).

\textbf{Image Classifier:} The basis of our saliency detection framework is a MobileNet~\cite{howard2017} binary classifier to categorize images either as healthy or as a sample with masses.
The network was trained on all training scans with batch size of 4 using the Adam optimizer with a learning rate ($lr$) of $1\text{e-}5$ for 250 epochs using early stopping.
Rotation, zoom, horizontal and vertical flips were used for data augmentation.
It was pretrained by 50k 224x224 pixel patches from the training data with the task of classifying background vs. masses.

\textbf{Inpainting:} The inpainter was trained on the healthy training samples with a batch size of 1 in two phases~\cite{liu2018}. The first phase was set up with batch normalization (BN) and $lr=1\text{e-}5$ for 100 epochs, the second without BN in the encoder part and with $lr=1\text{e-}6$ for 50 epochs. For each image up to 400 8x8 pixel holes were generated at random positions, where both single small holes and larger clusters were simulated to mimic configurations during optimization.

The inpainter has the task to change the classification score of an image towards healthy when replacing mass tissue, no considerable change should happen otherwise. 
To demonstrate that, we computed (i) a ROC curve using the classifier on all test samples without any inpainting, (ii) ROC curves for inpainting only in healthy tissue over 10 runs with randomly sampled holes and (iii) ROC curves for inpainting of mass tissue in unhealthy scans over 10 runs (Fig.~\ref{fig:inpainter_performance} left).

\textbf{Saliency Mapping:} Parametrization was experimentally chosen based on grid-search, restricted by $\lambda_i \in [0,1]$, for $i=1,2,3$. Similar to \cite{chang2018,fong2017} we found the resulting masks to be especially sensitive to $\lambda_2$. This smoothness controlling term, balances between noisy result-maps and compression induced information-loss. We exemplify this behaviour with an ablation study, where contributions of smoothing and sizing are set to zero (cf. Table ~\ref{tab:results_optimization}). The final optimization results were derived in 100 steps per image, with $lr=2\text{e-}3$, $\theta = 0.5$ and setting $\lambda_1 = 1.0$, $\lambda_{2,3} = 0.1$.

We compared our approach against two established methods based on widespread adaptation in medical imaging \cite{rajpurkar2017, baumgartner2017}, and inherent validity \cite{adebayo2018}. We chose the gradient based \emph{Saliency Map} \cite{simonyan2013} (SAL) and the network-derived \emph{Cam} \cite{zhou2016} (CAM) visualizations. As our domain prohibits the utilization of blurring, noise, etc. we could not meaningfully test against reference based methods~\cite{zintgraf2017,fong2017,dabkowski2017}. 
For evaluation, we derived four measures, that (i) relate the result masks (RM) to the mass annotations (GT), and (ii) compare the result masks to each other (cf. Table ~\ref{tab:results_optimization},~\ref{tab:results_overlap}). Particularly for (i) we studied $D$ the average of Euclidean distances between the centres of GT masks and RMs' connected components, and $H$ the average Hausdorff distances between GT masks and RM. Here, lower values indicate better localization, i.e. a better vicinity to the pathology.
Conversely, for (ii) we calculated $A$ the ratio between derived RMs and the organ masks (the breast area), and $O$ the overlap coefficients between our RMs and those of CAM and SAL. Here lower values of $A$ indicate a more compact map, i.e. better visual distinction. $O$ scores describe the correlation level between result masks. Statistically significant difference between all resulting findings was formalized using Wilcoxon signed-rank tests, for $\alpha < 0.01$.

All measurements were performed on binary masks, for which CAM and SAL had to be thresholded as those maps' noise covers the complete image. We therefore chose the $50/75/90$th percentiles, i.e. $50/25/10$ percent of the map-points. Where multiple masses, or mapping results occurred we used their median for a robust estimation per image. 
\section{Results and Conclusion}
\label{sec:results}

\textbf{Inpainting Evaluation:}
The ROC curves in Fig.~\ref{fig:inpainter_performance} represent an AUC of $0.89$ for original images (green), average AUCs of $0.88$ for inpainting tissue in healthy cases (blue) and $0.83$ for inpainting only in masses for pathological cases (red). Besides the AUCs, the visual separability of the green/blue curves from the red one indicates that the inpainter behaves correctly and introduces significant changes only when replacing mass tissue w.r.t. the classifier. The inpainting quality of replacing mass with healthy tissue is visible in Fig.~\ref{fig:inpainter_performance} right.

\textbf{Saliency Evaluation:}
\emph{Quantitatively}, our framework yields saliency masks significantly closer to GT masks based on Euclidean distances $D$ and Hausdorff distances $H$, 
both substantiated by p-values below $2\text{e-}12$ for all tested percentiles (cf. Table~\ref{tab:results_optimization}). 
Considering the map sizes $A$ (Table \ref{tab:results_overlap}), we report overall significantly smaller masks, and again p-values below $2\text{e-}12$ for all percentiles. This behavior changes when the shape-specific regularization parameters $\lambda_{2,3}$ are relaxed, as exemplified by the ablation study. As shown in the last row of both parts of Table~\ref{tab:results_optimization}, our feature attributions become scattered and noisy. Close inspection of the overlap-values, esp. $O$ in Table \ref{tab:results_overlap}, reveals that on average our method's attributions have a higher overlap with SAL than CAM. This indicates that our results tend to adhere to the dense localization spots of SAL, but alleviate the latter's noise and interpretation issue described in \cite{fong2017,zintgraf2017}.

\emph{Qualitatively}, as depicted in Fig.~\ref{fig:results} (b), our salient regions appear at the circumference of masses which is reasonable w.r.t. the fact that this is the discriminative region for the presence of masses. This is in line with~\cite{becker2019}, which reports on injection of poorly \textit{circumscribed}, malignant looking masses while transforming healthy cases into pathological ones using a GAN variant. 
In addition, our method yields more accurate visualizations than CAM and SAL (Fig.~\ref{fig:results} first row), i.e. it has
a smaller, more precise and more informative feature attribution than these standard visualization methods (Fig.~\ref{fig:results} (b)-(d)).

\textbf{Conclusion:}
We presented a novel, model agnostic, optimization-based, accurate saliency visualization approach tailored to medical images.
Image region marginalization was solved by partial convolution based inpainting, treating medical images correctly, thus overcoming the out-of-distribution problem.
Our method showed informative and faithful results on public mammography data, and is therefore suitable as a classification interpretation tool in radiological workflows.

\bibliographystyle{IEEEbib}
\bibliography{refs}

\end{document}